\pdfoutput=1

\documentclass{article}
\usepackage{colm2024_conference}

\usepackage{latexsym}

\usepackage[T1]{fontenc}

\usepackage[utf8]{inputenc}

\usepackage{microtype}

\usepackage{inconsolata}

\usepackage{graphicx}
\usepackage{ulem}
\usepackage{longtable}

\usepackage{url}
\usepackage{booktabs}
\usepackage{hyperref}
\usepackage{graphicx}
\usepackage{microtype}
\usepackage{multicol}
\usepackage{xcolor}   
\usepackage{multirow}
\usepackage{lipsum}
\usepackage{enumitem}
\usepackage{pifont}
\usepackage{colortbl}
\usepackage{tabularx}
\usepackage{caption} 
\usepackage{float}
\usepackage{amsmath,amsfonts,amssymb,bbm}
\usepackage{xspace}
\usepackage{amssymb}
\usepackage{cleveref}
\usepackage{fontawesome}
\usepackage{makecell}
\usepackage{marvosym}

\newcommand{\tabincell}[2]{\begin{tabular}{@{}#1@{}}#2\end{tabular}}

\newcommand{\bench}{\textsc{CodeElo}}

\begingroup

\footnotetext{Corresponding authors}
\endgroup

\title{
\bench{}: Benchmarking Competition-level Code Generation of LLMs with Human-comparable Elo Ratings
}

\author{
\textbf{Shanghaoran Quan \quad Jiaxi Yang \quad Bowen Yu \quad Bo Zheng \quad Dayiheng Liu \quad An Yang}\\
\textbf{Xuancheng Ren \quad Bofei Gao \quad Yibo Miao \quad Yunlong Feng \quad Zekun Wang} \\
\textbf{Jian Yang\quad Zeyu Cui \quad Yang Fan \quad Yichang Zhang \quad Binyuan Hui\textsuperscript{\Letter} \quad Junyang Lin\textsuperscript{\Letter}}\\
\vspace{0.25cm}
Qwen Team, Alibaba Group
\texttt{\{quanshanghaoran,binyuan.hby,junyang.ljy\}@alibaba-inc.com}
}

\begin{document}
\maketitle
\begin{abstract}
With the increasing code reasoning capabilities of existing large language models (LLMs) and breakthroughs in reasoning models like OpenAI o1 and o3, there is a growing need to develop more challenging and comprehensive benchmarks that effectively test their sophisticated competition-level coding abilities. 
Existing benchmarks, like LiveCodeBench and USACO, fall short due to the unavailability of private test cases, lack of support for special judges, and misaligned execution environments. 
To bridge this gap, we introduce \bench{}, a standardized competition-level code generation benchmark that effectively addresses all these challenges for the first time. 
\bench{} benchmark is mainly based on the official CodeForces\footnote{\url{https://codeforces.com}} platform and tries to align with the platform as much as possible. 
We compile the recent six months of contest problems on CodeForces with detailed information such as contest divisions, problem difficulty ratings, and problem algorithm tags. 
We introduce a unique judging method in which problems are submitted directly to the platform and develop a reliable Elo rating calculation system that aligns with the platform and is comparable with human participants but has lower variance. 
By testing on our \bench{}, we provide the Elo ratings of 30 existing popular open-source and 3 proprietary LLMs for the first time. 
The results show that o1-mini and QwQ-32B-Preview stand out significantly, achieving Elo ratings of 1578 and 1261, respectively, while other models struggle even with the easiest problems, placing in the lowest 25 percent among all human participants. 
Detailed analysis experiments are also conducted to provide insights into performance across algorithms and comparisons between using C++ and Python, which can suggest directions for future studies. 



\begin{figure*}[hb]
  \centering
  \hspace{-0.04\textwidth}
  \includegraphics[width=0.9\textwidth]{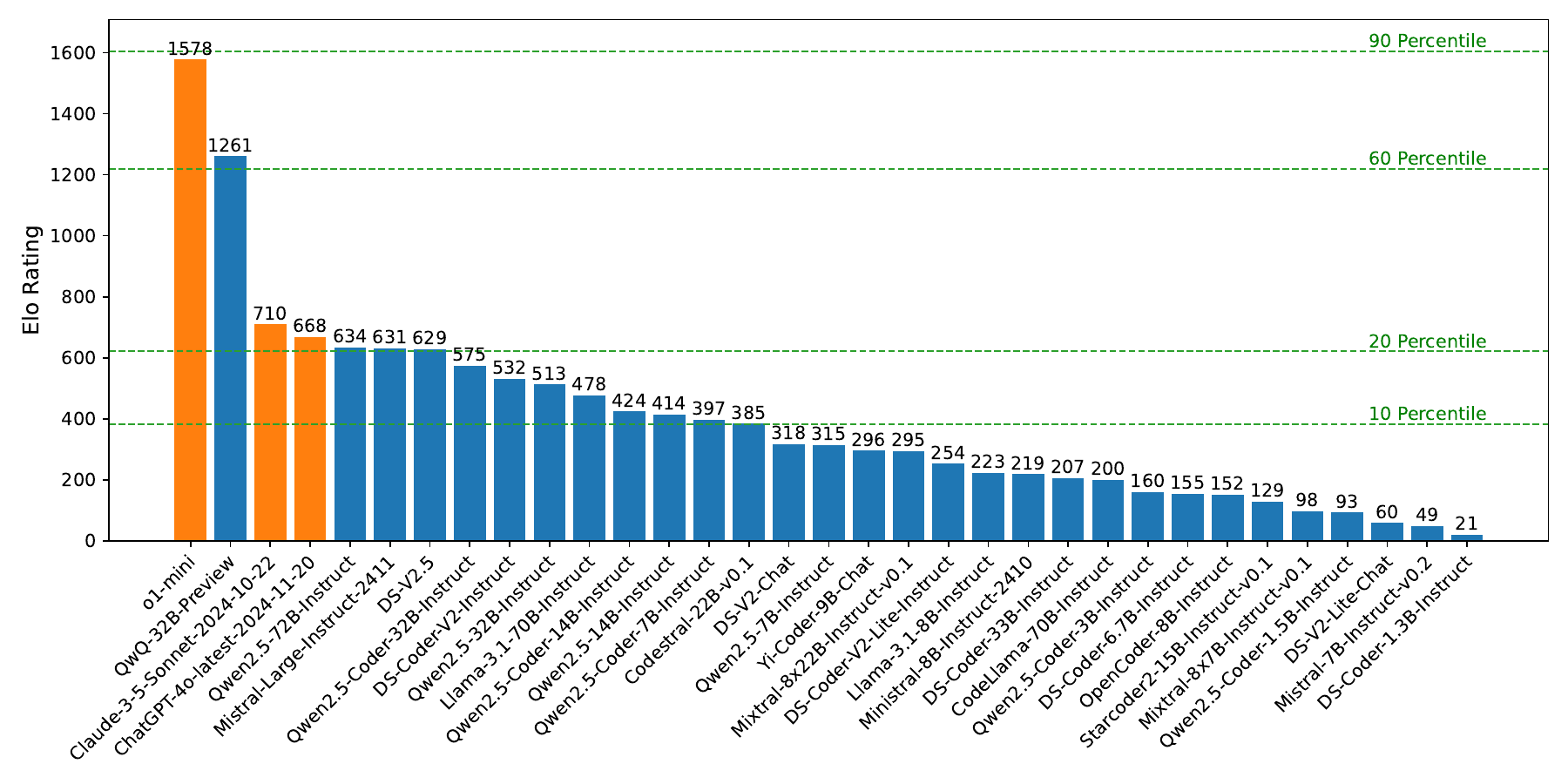}
  \vspace{-0.2cm}
  \caption{The Elo rating leaderboard. The test results may be slightly lower than the actual capabilities, as we constrain the number of submissions for each problem to eight times. The green dashed lines represent the Elo ratings of human participants at the corresponding percentiles.}
  \label{fig:ratings}
  \vspace{-0.2cm}
\end{figure*}

\end{abstract}

\section{Introduction}

With the increasing capabilities of existing LLMs and breakthroughs in reasoning models like OpenAI o1 and o3, there is a growing need to develop benchmarks that effectively test their sophisticated reasoning abilities. Math and coding are two evaluation methods for this purpose, as they provide accurate and easily verifiable feedback. While math presents hard benchmarks, like AIME~\citep{AIME2024}, Omni-MATH~\citep{gao2024omni}, and LiveAoPSBench~\citep{anonymous2024aops}, there's still a lack of a suitable benchmark in coding to appropriately measure LLMs' reasoning skills. We find that the CodeForces platform is suitable, and in the reports of OpenAI o1~\citep{o1-preview}, o3~\citep{o3} and DeepSeek r1~\citep{deepseek-r1}, they all test hard code on CodeForces. However, we still lack a standardized CodeForces test suite, leading to repeated compiling and creating work on CodeForces and potential misaligned settings and results. These situations highlight the need for research community to standardize a CodeForces evaluation benchmark to assess the competition-level coding capabilities of LLMs and reflect their corresponding sophisticated reasoning abilities.

We analyze that existing competition-level code benchmarks like LiveCodeBench~\citep{jain2024livecodebench}, USACO~\citep{shi2024can}, and CodeContests~\citep{li2022competition} 
cannot meet the above requirement due to certain limitations attributed to the following unique nature of competition code evaluation.
(1) Firstly, unlike typical math and general coding problems, competition code problem often requires extensive, human-designed, robust test cases to validate solution correctness. Although it is easy to scrape any problems from the websites, the competition code platforms or so-called online judges often hide their test cases. Consequently, existing benchmarks can only rely on publicly available or self-generated test cases that are often small and weak, leading to high false-positive rates in judgment\footnote{Passing all tests in CodeContests might still result in a wrong answer (\textasciitilde4\%) or a time limit error (\textasciitilde42\%), according to \url{https://alphacode.deepmind.com/}}. 
(2) Secondly, to truly evaluate LLM capabilities and provide human-comparable Elo ratings, it's necessary to test on all problems in contests just as human participants do. However, not all problems can be judged by directly comparing the output with the correct answer, as about 30\% of problems do not have unique correct outputs, which require specific judging codes called special judges\footnote{Special judges are programs used to determine whether solutions are accepted for problems that do not have a unique correct answer. A detailed discussion is provided in \Cref{appx:spj}.} to evaluate correctness. However, these special judges are often unavailable and difficult to write for some specific problems.
(3) Thirdly, different from general code testing, execution time is a significantly critical factor in competitive coding since problems often have strict constraints on running times and algorithm time complexity, but offline testing faces challenges due to varied efficiency across different machines, leading to potential misaligned evaluation results. Hence, there remains a lack of comprehensive standardized benchmarks for competition-level coding.

In this work, for the first time, we present the benchmark \bench{} and the corresponding evaluation method, which achieves zero false positives, supports special judges, and reaches an aligned execution environment to ensure absolutely correct judgments and provide human-comparable, standardized Elo ratings. We have compiled our test problems from CodeForces and categorized them by contest divisions, problem difficulty ratings, and problem algorithm tags for more comprehensive evaluation and analysis. To achieve absolutely correct judgments, we introduce a simple and effective method: using a bot to automatically submit model solutions to CodeForces and receive test results. Thanks to the judgment from the platform, this allows us to evaluate problems just like human participants, achieving zero false positives without needing access to the full test set, where test cases are often created adversarially and are robust. Similarly, given the favor of the platform judgments, we support special judges, which former benchmarks did not provide, and the execution environment is absolutely aligned since all use the same platform, even aligned with human participants. Based on the evaluation results and available user data on CodeForces, we also introduce an Elo rating calculation system that estimates the models' expected Elo rating based on their performance. This system is aligned with the CodeForces platform but has a much lower variance, as detailed in \Cref{sec:Elo-rating}.

We tested 30 existing popular open-source LLMs and 3 proprietary LLMs, with the leaderboard shown in \Cref{fig:ratings}. 
From the results, we find the OpenAI o1-mini model stands out by achieving the best performance with an Elo rating of 1578, surpassing nearly 90 percent of human participants. The o1-like reasoning model, QwQ-32B-Preview, also achieves great performance and stands out among open-source models, with an Elo rating of 1261, placing it in about the 60th percentile. This suggests that increasing the length of the chain-of-thought (CoT) is a promising way to improve the models' reasoning ability. 
On the other hand, most models struggle to pass the easiest problems and fall within the lowest 10th percentile of Elo ratings among human participants, and several large and prominent open-source LLMs fall in the range of the 10th to 20th percentiles.
Through more detailed experiments and analysis, we find that models are good at problems involving math and implementation, but struggle with dynamic programming (dp) and trees. We also find that for most models, the best-performing language is C++, instead of Python, which is most frequently used by LLMs and is also the main test language in previous benchmarks. We hope our \bench{} can pave the way for testing advanced LLMs' code reasoning capabilities and provide insights to improve such abilities in LLMs.

To summarize, our benchmark has the following main contributions, with a more detailed discussion on them available in \Cref{sec:contributions}.

\begin{itemize}[leftmargin=*]
\item We provide a set of Codeforces problems with detailed information like contest divisions, problem ratings, and problem algorithm tags.

\item We introduce a unique evaluation method in which problems are submitted directly to the platform, achieving zero false positives, special judge support and a fully aligned execution environment for the first time.

\item We are the first to provide standardized human-comparable Elo ratings that fairly judge the models' competition-level code generation for the existing popular open-source and proprietary LLMs.

\item A detailed analysis of experimental results provides insights that can suggest future studies.

\end{itemize}

\section{Related Work}

Before our work, there were several competitive code benchmarks. Here, we list six representative ones that are most relevant to our work:

\begin{itemize}[leftmargin=*]

\item \textbf{APPS}~\citep{hendrycks2021measuring}: Proposed in 2021/05, APPS curated problems from Codewars, AtCoder, Kattis, and CodeForces.

\item \textbf{CodeContests}~\citep{li2022competition}: Proposed in 2022/03, CodeContests includes problems, solutions, and test cases sourced from the CodeForces platform.

\item \textbf{xCodeEval}~\citep{khan2023xcodeeval}: Proposed in 2023/03, xCodeEval is a large-scale multilingual multitask code benchmark that also includes problems from CodeForces.

\item \textbf{TACO}~\citep{li2023taco}: Proposed in 2023/12, TACO compiles problems from CodeContests and APPS, adding new problems gathered from several competitive coding websites.

\item \textbf{LiveCodeBench}~\citep{jain2024livecodebench}: Proposed in 2024/03, LiveCodeBench primarily contains scraped coding problems and test cases from LeetCode and AtCoder, with minimal content from CodeForces. It avoids contamination by re-scraping new problems every month and releasing online updates.

\item \textbf{USACO}~\citep{shi2024can}: Proposed in 2024/04, USACO features hundreds of problems and test cases from the USA Computing Olympiad. It updates its benchmark through new released versions.

\end{itemize}

We include a comparison against these benchmarks in \Cref{tab:comparison}. We find that these benchmarks all source problems from open-access coding competition websites and conduct offline evaluations. While scraping problems is simple, most online judges hide their test cases. To address this, existing benchmarks attempt to generate test cases; however, these are often not as robust as the original ones (which are often designed in an adversarial manner and require a lot of labor from high-level participants), leading to many false-positive judgments. Additionally, these benchmarks do not support special judges. Another limitation is that they require execution on an individual's machine. Since runtime is a critical factor in algorithm competitions, differing machine performance can affect results. Furthermore, none of these benchmarks offer human-comparable standardized Elo ratings. These characteristics highlight the uniqueness and significant advantages of our benchmark.

\begin{table*}[ht]
  \centering
  \scalebox{0.95}{
    \begin{tabular}{lcccccc}
    \toprule
    & \tabincell{c}{\textbf{Problem} \\ \textbf{Diffculty}} & \textbf{Updates} &  \tabincell{c}{\textbf{Zero False} \\ \textbf{Positive?}} & \tabincell{c}{\textbf{Special} \\ \textbf{Judge?}} & \tabincell{c}{\textbf{Aligned} \\ \textbf{Execution} \\ \textbf{Environment?}} & \tabincell{c}{\textbf{Standardized} \\ \textbf{Elo Rating?}} \\
    \midrule
    APPS & $\bigstar\bigstar$ & No updates & \ding{55} & \ding{55}  & \ding{55} & \ding{55} \\
    CodeContests & $\bigstar\bigstar\bigstar$  & No updates & \ding{55} & \ding{55}  & \ding{55} & \ding{55} \\
    TACO & $\bigstar\bigstar$  & No updates & \ding{55} & \ding{55}  & \ding{55} & \ding{55} \\
    xCodeEval & $\bigstar$  & No updates & \ding{55} & \ding{55}  & \ding{55} & \ding{55} \\
    USACO & $\bigstar\bigstar$ & Offline & \ding{51} & \ding{55}  & \ding{55} & \ding{55} \\
    LiveCodeBench & $\bigstar$ & Online & \ding{55} & \ding{55}  & \ding{55} & \ding{55} \\
    \midrule
    \bench{} & $\bigstar\bigstar\bigstar$ & Online & \ding{51} & \ding{51} & \ding{51} & \ding{51} \\
    \bottomrule
    \end{tabular}%
  }
  \caption{Comparison between \bench{} and other competition code benchmarks. }
  \label{tab:comparison}%
\end{table*}%

Apart from competition code benchmarks, there are also some other more popular general code benchmarks, like HumanEval~\citep{chen2021evaluating}, MBPP~\citep{austin2021program} and BigCodeBench~\citep{zhuo2024bigcodebench}. These benchmarks have three significant differences with competition code benchmarks: 1) While we need to analyze the problem and write complete code with a focus on algorithm design, general benchmarks usually require writing only a small function to test specific functionality. 2) The problems in our test set are much harder than these general benchmarks, often needing sophisticated reasoning while general benchmarks do not. 3) Execution time is a significantly crucial constraint, as we not only judge the output for correctness but also demand on the algorithm complexity, whereas general benchmarks mainly focus on the former.



\section{\bench{} Benchmark}

Our benchmark is primarily based on the well-known coding competition platform, CodeForces. In this section, we discuss the development of our benchmark, detailing the processes of problem collection, classification and categorization, and the evaluation method.

\subsection{Problem Collection}

Our testing problems are sourced from the official CodeForces platform. We gathered all problems from the rated contests held since the platform inception (but we only tested on the recently held contests to avoid data contamination). 

An example problem can be found in Appendix \Cref{fig:prob-demo}. The scraped problems were originally in raw HTML format. We parsed out different sections of each problem, including the problem description, input format, output format, examples, notes, and so on, to allow for more flexible restructuring of the prompts during testing. By default, however, the problems are displayed in their original HTML format to preserve critical information and structure with minimal format translation for better clarity of text and formulas format. While these HTML formats might be challenging for humans to process without rendering, they can be easily processed by LLMs~\citep{gur2022understanding}.

\subsection{Classification and Categories}

The \bench{} includes several distinct classifications. These classifications are sourced and collected from the CodeForces platform, and we integrate them into our benchmark for more comprehensive analysis and detailed evaluations of various LLMs.

\paragraph{Contest Difficulty Division} CodeForces categorizes competitions into four divisions (Div. for short) based on their levels of difficulty: Div. 1, 2, 3, and 4, with Div. 1 being the most challenging and Div. 4 the least. Additionally, there are special divisions that combine Div. 1 and 2, as well as some global rounds whose difficulty averages between Div. 1 and 2. We classify these as Div. 1 + 2. It's important to note that each contest includes multiple problems, and the division just represents an overall average difficulty, so hard contests can also contain easier problems.

\paragraph{Problem Difficulty Rating} The problems also have a rating attribute but please note that this is a different concept with ratings for participants. The problem rating indicates the level of difficulty of a problem. Specifically, a problem rating of $x$ signifies that competitors with a rating of $x$ have a 50\% probability of passing the problem the first time they see it. The problem ratings are derived from actual human performance in contests and represent a statistically calculated value, making it relatively accurate.

\paragraph{Problem Algorithm Tags} Problems are categorized using algorithm tags, which identify the types of algorithms required to solve them. On average, a problem is associated with 3.9 tags, as it may need multiple steps or belong to different algorithmic categories. We have a total of 35 tags, and nearly 90\% of the occurrences are covered by the top 16 tags. Note that these tags are not visible to humans during contests and also not visible to the models being tested, thus they cannot serve as hints for solutions.

\subsection{Evaluation Method}

\subsubsection{Solution Submitting and Judgment}

Unlike other benchmarks that require testing on one's own machine, which may cause execution environment misalignment, we parse the solution code block and directly submit it to the CodeForces official platform after obtaining the model's output to achieve absolutely accurate feedback. This is especially important in competitive coding problems since execution time is a crucial metric in judging a solution, and different environments may vary in this aspect. The proxy judging settings are also advantageous because they not only allow us to bypass the need for complete test cases but also support the evaluation of problems that require special judges, thereby enabling us to better assess the model's capabilities and provide more accurate results.

We use an automatic submission bot to help submit the problems and obtain judging results. Since the results come directly from the official platform, we can directly parse the status to see if it is an "Accept." Unlike some other competitive coding benchmarks like APPS, we consider a solution accepted only when it passes all the test cases; partially passing a proportion of test cases does not count towards scoring, aligned with the criteria on the platform.

\subsubsection{Elo Rating Calculation System}
\label{sec:Elo-rating}

We use an Elo rating calculation method similar to the official CodeForces platform Elo rating calculation system\footnote{\url{https://codeforces.com/blog/entry/20762}} to obtain a standardized Elo rating. This rating reflects an individual's competitive programming ability and is comparable between humans, models, and across both humans and models. Unlike CodeForces, which updates a participant's rating by considering changes after each contest to maintain it online, we treat each contest as independent for simplicity and accuracy. Thus, we calculate the model's expected rating \(r\) for each contest individually.

Specifically, let's assume there are \(n\) human participants in a contest with ratings \(r_{(i)}\) for \(i = 1, 2, ..., n\). For ease of mathematical representation, assume participants are ranked from best to worst in terms of performance in this contest. Position the model's performance within this ranking, denoting the model's rank as \(m\) (where \(1 \leq m \leq n+1\)). Suppose the model's expected rating is \(r\). According to the definition of the Elo rating \citep{elo1978rating}, we have the following equation:

\[
    m = \sum\limits_{i=1}^{n} \frac{1}{1 + 10^{(r - r_{(i)})/400}}
\]

Since the expression on the right side of the equation is monotonically decreasing with respect to \(r\), we can easily use binary search to determine the exact value of \(r\).

As demonstrated in \Cref{appx:Elo-rating}, we have analyzed that this calculation method maintains the same expected rating as the official platform's method but with significantly lower variance.

\paragraph{Advantages of Elo Ratings Over Other Metrics} While many existing benchmarks use pass@$n$ (with $n$ typically being $1$ or other values) as the evaluation metric, our benchmark adopts a more advanced Elo rating system. This system accounts for multiple attempts, providing a more comprehensive analysis than simply considering pass@$1$ and takes the sampling diversity into account. 
Moreover, it effectively balances pass attempts by penalizing failed attempts before the successful ones, which is superior to the traditional pass@$n$ metric.
In addition, passing more challenging problems will receive higher scores in rating calculations, encouraging models to tackle more difficult tasks, which is also a feature not supported by most previous benchmarks.

\section{Evaluation on Existing LLMs}

\subsection{Experiment Setup}

Based on our observation that most models struggle with even the simplest problems in Div. 1 contests, we decided to discard these contests and focus solely on testing the remaining divisions in the most recent contests. We gathered contests held between May 4, 2024, and November 4, 2024, totaling 54 contests or 387 problems. We present basic statistics for these contests in \Cref{tab:contest-div}.

\begin{table}[h!]
    \centering
    \begin{tabular}{>{\bfseries}c|cccc}
        \toprule
        Div. & \textbf{Count} & \tabincell{c}{\textbf{Avg.} \\ \textbf{Problems}} & \tabincell{c}{\textbf{Avg.} \\ \textbf{Ratings}} & \tabincell{c}{\textbf{Rating} \\ \textbf{Requirement}} \\
        \midrule
        \sout{1}     & \sout{3} & \sout{7.0} & \sout{2533} & \sout{$\geq$ 1900}  \\
        1+2   & 8 & 9.1 & 2106 & $All$ \\
        2     & 33 & 6.5 & 1779 & $\le$ 2100   \\
        3     & 10 & 7.5 & 1436 & $\le$ 1600   \\
        4     & 3 & 8.3 & 1276 & $\le$ 1400   \\
        \bottomrule
    \end{tabular}
    \caption{Basic statics of different contest divisions.}
    \label{tab:contest-div}
\end{table}

For each problem, we allow each tested model up to eight attempts. Given that the inference times for each model are significantly lower than those for humans and can be omitted, we assume that models will respond and submit solutions within the first minute, so no time penalties are applied. However, penalty points will still be counted for any failed attempts made before a successful submission, which aligned with the platform. Each problem has a specific score, and we calculate the final scores and penalties of the tested models to compare and rank them against human participants. All conditions not mentioned are kept as similar as possible to the official platform settings.

We evaluated 30 popular open-source models using vLLM for inference and 4 proprietary models via API calls (detailed in the model cards found in \Cref{appx:model-card}). All tested models followed the same chain-of-thought prompting: 
\begin{quote}
    You are a coding expert. Given a competition-level coding problem, you need to write a C++ program to solve it. You may start by outlining your thought process. In the end, please provide the complete code in a code block enclosed with \`{}\`{}\`{} \`{}\`{}\`{}.
\end{quote}
Here C++ was chosen as the test language because it generally elicits the best performance from models on competition code problems. A detailed analysis of language use is provided in \Cref{sec:test-lang}.

\subsection{Elo Ratings}

We present the standardized Elo rating leaderboard in \Cref{fig:ratings}. Our analysis reveals that o1-mini, with a rating of 1578, and QwQ-32B-Preview, with a rating of 1261, stand out significantly among proprietary and open-source models, respectively. This demonstrates the substantial advantage of long CoT o1-like reasoning models in tackling difficult competition code problems. Additionally, we observe a clear trend that larger models tend to outperform smaller ones, and all open-source models that achieved a rating above 500 are 30B+ models. However, their ratings remain below 700, which corresponds to approximately the lowest 20th percentile among all human participants in \Cref{tab:percentile-rating}.

\subsection{Main Results}

We present performance details of all tested proprietary and open-source models in \Cref{tab:main-res}. For a clearer comparison, we categorize open-source models by their sizes and classify mixture-of-experts models by regarding their parameters as square root of product of activation and total parameters.

\begin{table*}[!ht]
 \centering
 \resizebox{1.0\textwidth}{!}{
 \begin{tabular}{lr|ccccc|ccc|cccc}
 \toprule
\multirow{2}{*}{\textbf{Model}} &  & \multicolumn{5}{c|}{$\textbf{Elo Rating}$}
                                & \multicolumn{3}{c|}{\textbf{Pass Rate for}}
                                & \multicolumn{4}{c}{\textbf{Pass @}} \\
\cmidrule(lr){3-7} \cmidrule(lr){8-10}  \cmidrule(lr){11-14}
      & 
      & Overall
      & Div. 1 + 2
      & Div. 2
      & Div. 3
      & Div. 4
      & Easy
      & Medium
      & Hard
      & 1
      & 2
      & 4
      & 8 \\\midrule
 \multicolumn{14}{c}{\textbf{Proprietary LLMs}} \\ \midrule

ChatGPT-4o-latest-2024-11-20 & \faLock{} & 668 (22.2) & 586 & 507 & 1111 & 1149 & 36.54 & 14.0 & 0.83 & 9.3 & 10.8 & 14.57 & 16.83 \\
Claude-3-5-Sonnet-2024-10-22 & \faLock{} & 710 (24.1) & 430 & 616 & 1092 & 1124 & 46.47 & 11.0 & 0.97 & 11.81 & 13.82 & 15.58 & 16.08 \\
o1-mini & \faLock{} & \underline{1578} (89.2) & \underline{1197} & \underline{1541} & \underline{1906} & \underline{1792} & \underline{74.52} & \underline{42.75} & \underline{11.71} & \underline{26.88} & \underline{33.92} & \underline{39.7} & \underline{39.95} \\

\midrule
 \multicolumn{14}{c}{\textbf{1B+ Open-source LLMs}} \\ \midrule
 
DS-Coder-1.3B-Instruct & 1.3B & 21 (0.0) & 0 & 0 & 0 & 378 & 3.37 & 0.0 & 0.0 & 0.75 & 0.75 & 0.75 & 0.75 \\
Qwen2.5-Coder-1.5B-Instruct & 1.5B & 93 (0.0) & 0 & 48 & 179 & 514 & 6.73 & 0.0 & 0.0 & 1.26 & 1.76 & 2.51 & 2.51 \\
Qwen2.5-Coder-3B-Instruct & 3B & \underline{160} (0.0) & 0 & \underline{74} & \underline{398} & \underline{676} & \underline{10.9} & \underline{0.5} & 0.0 & \underline{2.26} & \underline{2.51} & \underline{4.02} & \underline{4.77} \\

 \midrule
 \multicolumn{14}{c}{\textbf{6B+ Open-source LLMs}} \\ \midrule
 
Mistral-7B-Instruct-v0.2 & 7B & 49 (0.0) & 0 & 0 & 146 & 378 & 6.25 & 0.0 & 0.0 & 1.26 & 1.26 & 1.26 & 1.26 \\
DS-V2-Lite-Chat & 2.4/16B & 60 (0.0) & 0 & 16 & 151 & 378 & 4.01 & 0.0 & 0.0 & 1.01 & 1.26 & 1.76 & 1.76 \\
OpenCoder-8B-Instruct & 8B & 152 (0.0) & 0 & 70 & 372 & 667 & 8.17 & 0.5 & 0.0 & 1.01 & 2.51 & 3.77 & 4.52 \\
DS-Coder-6.7B-Instruct & 6.7B & 155 (0.0) & 0 & 97 & 319 & 606 & 10.1 & 0.25 & 0.0 & 1.76 & 2.26 & 3.27 & 4.52 \\
Ministral-8B-Instruct-2410 & 8B & 219 (0.0) & 0 & 118 & 548 & 745 & 13.94 & 0.5 & 0.05 & 2.51 & 3.52 & 4.77 & 6.28 \\
Llama-3.1-8B-Instruct & 8B & 223 (0.0) & 0 & 207 & 325 & 585 & 12.18 & 0.25 & 0.0 & 2.26 & 2.76 & 4.52 & 6.53 \\
DS-V2-Lite-Instruct & 2.4/16B & 254 (0.0) & \underline{187} & 155 & 446 & \underline{851} & 16.51 & \underline{3.5} & 0.05 & 3.02 & 4.27 & 5.78 & 6.78 \\
Yi-Coder-9B-Chat & 9B & 296 (0.0) & 108 & 228 & 560 & 606 & 14.26 & 1.75 & 0.09 & 2.76 & 4.02 & 6.28 & 7.29 \\
Qwen2.5-7B-Instruct & 7B & 315 (0.0) & 123 & 242 & 581 & 676 & 17.63 & 1.5 & 0.09 & 4.27 & 5.53 & 6.78 & 7.79 \\
Qwen2.5-Coder-7B-Instruct & 7B & \underline{397} (11.6) & 143 & \underline{334} & \underline{647} & 842 & \underline{19.55} & 3.0 & \underline{0.14} & \underline{4.52} & \underline{6.03} & \underline{8.29} & \underline{10.05} \\

 \midrule
 \multicolumn{14}{c}{\textbf{13B+ Open-source Models}} \\ \midrule
 
Mixtral-8x7B-Instruct-v0.1 & 8/56B & 98 (0.0) & 0 & 30 & 226 & 644 & 5.29 & 0.25 & 0.05 & 1.26 & 1.51 & 2.26 & 3.52 \\
Starcoder2-15B-Instruct-v0.1 & 15B & 129 (0.0) & 0 & 48 & 333 & 641 & 5.93 & 0.0 & 0.0 & 1.76 & 2.76 & 3.27 & 3.52 \\
Codestral-22B-v0.1 & 22B & 385 (10.2) & 57 & \underline{345} & 586 & 926 & 20.03 & 2.25 & 0.14 & 3.52 & 4.77 & 6.78 & 10.3 \\
Qwen2.5-14B-Instruct & 14B & 414 (12.9) & \underline{497} & 277 & 752 & 606 & 23.4 & 1.5 & \underline{0.32} & 5.03 & 6.03 & 7.79 & 11.31 \\
Qwen2.5-Coder-14B-Instruct & 14B & \underline{424} (13.5) & 123 & 292 & \underline{876} & \underline{1067} & \underline{25.64} & \underline{5.75} & \underline{0.32} & \underline{6.78} & \underline{8.04} & \underline{9.55} & \underline{12.06} \\

\midrule 
 \multicolumn{14}{c}{\textbf{30B+ Open-source Models}} \\ \midrule
 
CodeLlama-70B-Instruct & 70B & 200 (0.0) & 0 & 111 & 539 & 507 & 8.97 & 0.75 & 0.05 & 1.76 & 2.76 & 5.03 & 5.78 \\
DS-Coder-33B-Instruct & 33B & 207 (0.0) & 0 & 113 & 498 & 746 & 13.46 & 1.5 & 0.0 & 3.02 & 4.27 & 4.52 & 6.28 \\
Mixtral-8x22B-Instruct-v0.1 & 22/176B & 295 (0.0) & 75 & 241 & 510 & 680 & 14.42 & 0.5 & 0.05 & 3.27 & 4.27 & 5.78 & 7.04 \\
DS-V2-Chat & 21/236B & 318 (0.0) & 59 & 253 & 588 & 737 & 16.83 & 2.25 & 0.0 & 3.77 & 4.77 & 6.53 & 9.05 \\
Llama-3.1-70B-Instruct & 70B & 478 (15.0) & 255 & 361 & 886 & 933 & 25.32 & 3.0 & 0.46 & 5.03 & 7.29 & 10.55 & 12.56 \\
Qwen2.5-32B-Instruct & 32B & 513 (15.5) & 350 & 366 & 923 & 1146 & 28.85 & 6.5 & 0.46 & 5.53 & 8.29 & 10.8 & 13.07 \\
DS-Coder-V2-Instruct & 21/236B & 532 (15.7) & 420 & 400 & 853 & 1179 & 29.33 & 7.5 & 0.37 & 6.53 & 8.79 & 11.81 & 14.32 \\
Qwen2.5-Coder-32B-Instruct & 32B & 575 (16.8) & 206 & 416 & 1166 & 1222 & 29.49 & 7.75 & 0.46 & 6.03 & 8.54 & 12.06 & 16.58 \\
DS-V2.5 & 21/236B & 629 (20.4) & 594 & 483 & 958 & 1221 & 33.65 & 10.0 & 0.65 & 8.79 & 11.56 & 13.32 & 15.58 \\
Mistral-Large-Instruct-2411 & 123B & 631 (20.5) & 632 & 449 & 1049 & 1226 & 35.58 & 9.5 & 0.65 & 8.29 & 11.81 & 13.07 & 16.33 \\
Qwen2.5-72B-Instruct & 72B & 634 (20.7) & 439 & 498 & 1033 & 1255 & 35.26 & 12.0 & 0.97 & 9.3 & 11.06 & 13.32 & 16.58 \\
QwQ-32B-Preview & 32B & \underline{1261} (63.6) & \underline{1071} & \underline{1169} & \underline{1566} & \underline{1700} & \underline{57.21} & \underline{21.75} & \underline{4.54} & \underline{18.59} & \underline{23.12} & \underline{29.4} & \underline{32.91} \\

 \bottomrule
 \end{tabular}
 }
 \caption{Main results of different LLMs on \bench{}. The number in parentheses after the overall Elo rating shows the percentile rank among human participants. The underlined numbers represent the best scores within the same model size range.}
 \label{tab:main-res}
 
\end{table*}

\paragraph{Elo Ratings Across Contest Divisions} From a contest perspective, we observe that contests in easier divisions often result in higher performance ratings. However, we note that this trend may not always hold true if the model's capacity is larger. The best results are typically achieved when models are tested in contests that match their skill level (refer to rating requirements in \Cref{tab:contest-div} for approximate contest ratings based on Elo ratings). Consequently, most models perform optimally in Div. 4. However, o1-mini demonstrates the best performance in Div. 3, consistent with its higher rating and the rating requirements for that division. From a model perspective, we find that superior models consistently outperform inferior ones across different contest divisions.

\paragraph{Pass Rate Across Problem Difficulty Levels}

We flatten the problems in contests and divide them by problem ratings: Easy ($[800, 1000)$), Medium ($[1000, 1300)$), and Hard ($[1300, 3500]$). We record the model pass rate at these difficulty levels and find that even the easy category is challenging for most models and demonstrates the best differentiation. The medium level effectively distinguishes several advanced models. Although the minimum problem rating in the hard level is only 1300, all models, except for o1-mini and QwQ-32B-Preview, are struggling to achieve a pass. These results also indicate significant room for improvement in existing models and highlight the extensibility of our benchmark.

\paragraph{Pass@$\mathbf{n}$} We also display the pass@$n$ metrics for different values of $n (n = 1, 2, 4, 8)$ for each model. Our findings indicate that most models increase their pass rate consistently as the number of samplings rises. Some models may have poor pass@$1$ results but improve significantly by pass@$8$, showcasing the model's ability to produce diverse effective explorations. Moreover, we find that pass@$n$ often correlates with Elo ratings, but they may not always align due to their differing calculation methods.

\section{Analysis Experiments}

\subsection{Performance Across Algorithms}

Each problem is associated with tags that suggest potential algorithms or methods for solving it. We have identified 16 tags that appear frequently, each corresponding to at least 30 problems tested. Note that a single problem may be associated with multiple tags, as it may require multiple steps or fall under different algorithmic categories. The performance of different models on these tagged problems is summarized in \Cref{tab:algo-tags}.

\begin{table*}[!ht]
 \centering
 \resizebox{1.0\textwidth}{!}{
 \begin{tabular}{lr|cccccccccccccccc}
 \toprule
\textbf{Model} &  & Gr. & Ma. & Im. & BF. & DP & DS. & CA. & BS. & So. & Gr. & DFS & NT. & Tr. & Co. & TP. & Bi. \\ \midrule

\multicolumn{18}{c}{\textbf{Proprietary LLMs}} \\ \midrule

ChatGPT-4o-latest-2024-11-20 & \faLock{} & 5.60 & 9.07 & 12.80 & 9.53 & 2.17 & 1.59 & 6.39 & 4.17 & 14.58 & 1.82 & 0.00 & 4.83 & 0.00 & 4.28 & 6.07 & 2.57 \\
Claude-3-5-Sonnet-2024-10-22 & \faLock{} & 9.40 & 12.02 & 15.97 & 10.35 & 0.00 & 2.50 & 7.07 & 5.25 & 17.50 & 0.78 & 0.80 & 5.11 & 0.00 & 3.62 & 7.50 & 2.94 \\
o1-mini & \faLock{} & \underline{25.83} & \underline{31.11} & \underline{31.94} & \underline{23.98} & \underline{10.65} & \underline{14.77} & \underline{22.15} & \underline{22.38} & \underline{34.58} & \underline{13.54} & \underline{11.44} & \underline{22.73} & \underline{4.55} & \underline{25.00} & \underline{19.29} & \underline{20.59} \\

\midrule
 \multicolumn{18}{c}{\textbf{1B+ Open-source LLMs}} \\ \midrule

DS-Coder-1.3B-Instruct & 1.3B & 0.00 & 0.55 & 2.08 & 0.00 & 0.00 & 0.00 & 0.00 & 0.00 & 1.46 & 0.00 & 0.00 & 0.00 & 0.00 & 0.00 & 0.00 & 0.00 \\
Qwen2.5-Coder-1.5B-Instruct & 1.5B & 0.06 & 1.10 & 3.27 & 0.51 & 0.00 & 0.00 & 0.00 & 0.62 & 2.08 & 0.00 & 0.00 & 0.57 & 0.00 & 0.00 & 0.00 & 0.00 \\
Qwen2.5-Coder-3B-Instruct & 3B & \underline{0.54} & \underline{2.06} & \underline{3.97} & \underline{1.74} & 0.00 & \underline{0.23} & \underline{0.68} & \underline{0.77} & \underline{3.96} & 0.00 & 0.00 & \underline{1.42} & 0.00 & 0.00 & \underline{1.07} & 0.00 \\

\midrule
 \multicolumn{18}{c}{\textbf{6B+ Open-source LLMs}} \\ \midrule

Mistral-7B-Instruct-v0.2 & 7B & 0.00 & 1.03 & 3.17 & 0.72 & 0.00 & 0.00 & 0.00 & 0.00 & 3.12 & 0.00 & 0.00 & 0.00 & 0.00 & 0.00 & 0.00 & 0.00 \\
DS-V2-Lite-Chat & 2.4/16B & 0.06 & 0.69 & 2.28 & 0.20 & 0.00 & 0.00 & 0.14 & 0.00 & 1.67 & 0.00 & 0.00 & 0.00 & 0.00 & 0.00 & 0.00 & 0.00 \\
OpenCoder-8B-Instruct & 8B & 0.54 & 1.24 & 4.07 & 0.51 & 0.00 & 0.11 & 0.95 & 0.15 & 2.29 & 0.00 & 0.00 & 0.28 & 0.00 & 0.33 & 0.36 & 0.00 \\
DS-Coder-6.7B-Instruct & 6.7B & 0.60 & 1.79 & 4.17 & 1.23 & 0.00 & 0.00 & 0.54 & 0.62 & 2.92 & 0.00 & 0.00 & 1.42 & 0.00 & 0.33 & 0.36 & 0.00 \\
Ministral-8B-Instruct-2410 & 8B & 1.01 & 2.40 & 5.36 & 1.84 & 0.00 & 0.34 & 0.27 & 1.08 & 3.96 & 0.00 & 0.00 & 1.99 & 0.00 & 0.00 & 0.71 & 0.00 \\
Llama-3.1-8B-Instruct & 8B & 0.65 & 2.61 & 4.76 & 1.13 & 0.00 & 0.00 & 1.49 & 0.62 & 3.54 & 0.00 & 0.00 & 1.14 & 0.00 & 0.33 & 0.36 & 0.00 \\
DS-V2-Lite-Instruct & 2.4/16B & 1.73 & 3.78 & \underline{6.85} & \underline{3.48} & 0.11 & 0.68 & 1.77 & 1.08 & 5.21 & 0.00 & 0.00 & \underline{2.27} & 0.00 & \underline{1.97} & 1.43 & 0.00 \\
Yi-Coder-9B-Chat & 9B & 1.67 & 2.82 & 5.85 & 2.15 & \underline{0.43} & 0.23 & 1.90 & 1.23 & 5.62 & 0.00 & 0.00 & 1.42 & 0.00 & 0.33 & 0.00 & \underline{0.37} \\
Qwen2.5-7B-Instruct & 7B & 1.49 & 3.78 & 5.36 & 2.97 & 0.00 & 0.80 & 1.49 & 1.54 & 5.00 & \underline{0.26} & \underline{0.27} & \underline{2.27} & 0.00 & 0.00 & \underline{2.14} & 0.00 \\
Qwen2.5-Coder-7B-Instruct & 7B & \underline{2.14} & \underline{3.98} & 6.55 & 3.38 & 0.11 & \underline{1.02} & \underline{2.04} & \underline{1.85} & \underline{7.29} & 0.00 & 0.00 & \underline{2.27} & 0.00 & 0.33 & 1.07 & \underline{0.37} \\

\midrule
\multicolumn{18}{c}{\textbf{13B+ Open-source Models}} \\ \midrule

Mixtral-8x7B-Instruct-v0.1 & 8/56B & 0.06 & 1.17 & 2.18 & 0.51 & 0.00 & 0.00 & 0.27 & 0.93 & 1.25 & 0.00 & 0.00 & 0.57 & 0.00 & 0.00 & 0.00 & 0.00 \\
Starcoder2-15B-Instruct-v0.1 & 15B & 0.36 & 0.96 & 2.78 & 0.61 & 0.00 & 0.00 & 0.68 & 0.00 & 2.29 & 0.00 & 0.00 & 0.57 & 0.00 & 0.00 & 0.00 & 0.00 \\
Codestral-22B-v0.1 & 22B & 2.32 & 3.71 & 9.03 & 2.77 & 0.00 & 0.23 & \underline{2.45} & 0.77 & 6.46 & 0.00 & \underline{0.27} & 1.70 & \underline{0.28} & 0.33 & 0.36 & \underline{0.37} \\
Qwen2.5-14B-Instruct & 14B & 3.21 & 5.43 & 7.94 & 3.38 & 0.65 & 1.14 & 2.31 & 2.01 & 7.29 & 0.26 & \underline{0.27} & 2.56 & 0.00 & 0.66 & 1.43 & \underline{0.37} \\
Qwen2.5-Coder-14B-Instruct & 14B & \underline{3.33} & \underline{5.63} & \underline{9.13} & \underline{5.74} & \underline{1.20} & \underline{1.36} & \underline{2.45} & \underline{2.47} & \underline{9.58} & \underline{1.04} & \underline{0.27} & \underline{2.84} & 0.00 & \underline{0.99} & \underline{2.86} & 0.00 \\

\midrule 
 \multicolumn{18}{c}{\textbf{30B+ Open-source Models}} \\ \midrule

CodeLlama-70B-Instruct & 70B & 0.48 & 1.65 & 3.87 & 0.92 & 0.00 & 0.34 & 0.41 & 0.62 & 2.92 & 0.00 & 0.00 & 0.85 & 0.00 & 0.66 & 0.71 & 0.00 \\
DS-Coder-33B-Instruct & 33B & 1.37 & 2.40 & 5.36 & 1.33 & 0.33 & 0.11 & 0.82 & 0.77 & 3.54 & 0.00 & 0.00 & 1.99 & 0.00 & 0.00 & 0.71 & 0.37 \\
Mixtral-8x22B-Instruct-v0.1 & 22/176B & 1.55 & 3.09 & 5.56 & 1.95 & 0.00 & 0.11 & 1.90 & 1.08 & 7.29 & 0.00 & 0.00 & 0.85 & 0.00 & 0.00 & 0.71 & 0.37 \\
DS-V2-Chat & 21/236B & 1.61 & 3.57 & 6.35 & 2.77 & 0.11 & 0.68 & 1.90 & 1.23 & 4.58 & 0.00 & 0.00 & 2.27 & 0.00 & 0.00 & 0.00 & 0.00 \\
Llama-3.1-70B-Instruct & 70B & 2.98 & 5.98 & 10.02 & 4.00 & 0.33 & 0.80 & 2.72 & 2.78 & 9.17 & 0.00 & 0.00 & 2.27 & 0.00 & 0.66 & 2.86 & 1.10 \\
Qwen2.5-32B-Instruct & 32B & 3.75 & 6.59 & 9.72 & 4.51 & 0.87 & 1.59 & 3.67 & 3.86 & 10.83 & 0.78 & 0.00 & 2.84 & 0.00 & 0.66 & 1.79 & 1.47 \\
DS-Coder-V2-Instruct & 21/236B & 3.81 & 6.94 & 11.21 & 4.82 & 1.09 & 1.14 & 3.94 & 2.93 & 8.75 & 2.08 & 0.00 & 2.84 & 0.00 & 2.63 & 5.00 & 1.10 \\
Qwen2.5-Coder-32B-Instruct & 32B & 4.05 & 7.01 & 9.62 & 6.35 & 1.52 & 1.59 & 4.76 & 4.01 & 11.04 & 1.30 & 0.27 & 3.41 & 0.00 & 1.32 & 5.00 & 0.74 \\
DS-V2.5 & 21/236B &  \hspace{0.3mm} 5.18 & 8.24 & 13.10 & 6.05 & 1.30 & 1.70 & 4.89 & 2.62 & 12.71 & 2.08 & 0.00 & 2.56 & 0.00 & 3.62 & 3.57 & 1.84 \\
Mistral-Large-Instruct-2411 & 123B & 6.01 & 8.17 & 11.61 & 6.05 & 1.63 & 2.16 & 4.48 & 4.78 & 13.96 & 1.04 & 0.00 & 2.84 & 0.00 & 0.99 & \underline{7.50} & 2.94 \\
Qwen2.5-72B-Instruct & 72B & 6.79 & 9.00 & 12.40 & 7.68 & 1.41 & 1.48 & 7.34 & 3.70 & 16.88 & 2.60 & 1.33 & 3.12 & 0.00 & 1.32 & 5.71 & 1.84 \\
QwQ-32B-Preview & 32B & \underline{15.00} & \underline{21.70} & \underline{19.64} & \underline{15.37} & \underline{3.37} & \underline{8.18} & \underline{15.35} & \underline{8.80} & \underline{23.96} & \underline{4.17} & \underline{3.19} & \underline{9.66} & \underline{0.57} & \underline{14.14} & 6.43 & \underline{8.09} \\

 \bottomrule
 \end{tabular}
 }
 \caption{Pass rate (pass@$1$) on major problem categories that have at least 30 problems tested. The abbreviations "Gr.", "Ma.", "Im.", "BF.", "DP", "DS.", "CA.", "BS.", "So.", "Gr.", "DFS", "NT.", "Tr.", "Co.", "TP.", and "Bi." stand for greedy, math, implementation, brute force, dp, data structures, constructive algorithms, binary search, sortings, graphs, dfs and similar, number theory, trees, combinatorics, two pointers, and bitmasks, respectively.}
 \label{tab:algo-tags}
\end{table*}

We observe significant variation in model performance across different algorithms. Models demonstrate strong performance in areas such as math (Ma.), implementation (Im.), and sorting (So.), achieving the highest pass rates. However, they struggle with dp (DP), dfs and similar (DFS), and trees (Tr.), with many models failing to solve even a single problem under these algorithms.

\subsection{Comparison between C++ and Python}
\label{sec:test-lang}

An interesting observation is that while Python is the most commonly used programming language for most models, the best performance on \bench{} is achieved when models use C++ as the coding language. This performance surpasses even the scenarios where models select the language on their own.

We conducted experiments where models received prompts without a specified programming language, allowing them to choose freely. We found that under our \bench{}, nearly all models defaulted to using Python over 95\% of the time, with only occasional use of C++, Java, and others. In contrast, human participants in code competitions predominantly use C++, with rates close to 80\%\footnote{We randomly selected 250 human submissions and found 201 used C++, and this ratio will be even higher among proficient competitors.}. We further investigated the performance difference of models when constrained to use either C++ or Python.

\begin{figure}[!h]
  \centering
  \includegraphics[width=0.5\linewidth]{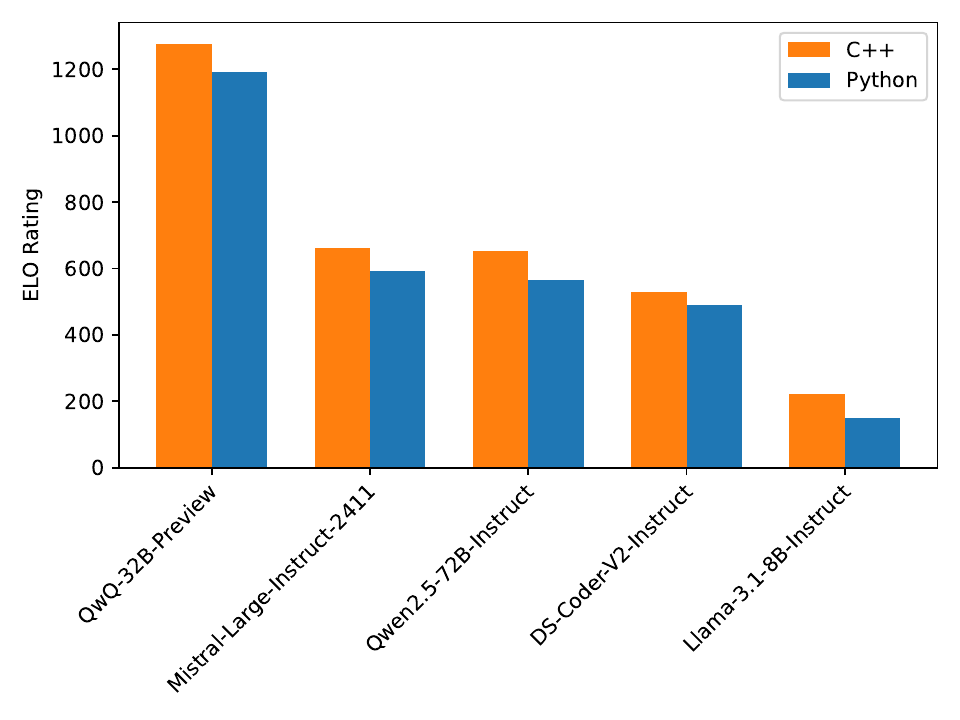}
  \caption{The Elo ratings using C++ and Python as programming languages.}
  \label{fig:lang-rating}
  \vspace{-0.4cm}
\end{figure}

We instructed several popular LLMs by specifying the coding language in prompts, and the results are displayed in \Cref{fig:lang-rating}. Our findings show that all models achieved higher ratings when using C++. This aligns with human experience, as competition-level code problems often impose constraints on algorithm running time, and C++ is more efficient at meeting these challenges. These findings disclose existing model training shortcomings and offer guidance on enhancing model language selection cognition: models should be trained to use C++ when facing problems where runtime efficiency is critical to get better performance. These findings also indicate that we had better test the models in C++ to unlock their best performance. This insight contrasts existing competition-level code benchmarks like APPS and LiveCodeBench, which predominantly assess performance using Python.

\subsection{Rating Variance}

\begin{figure}[!h]
  \centering
  \includegraphics[width=0.5\linewidth]{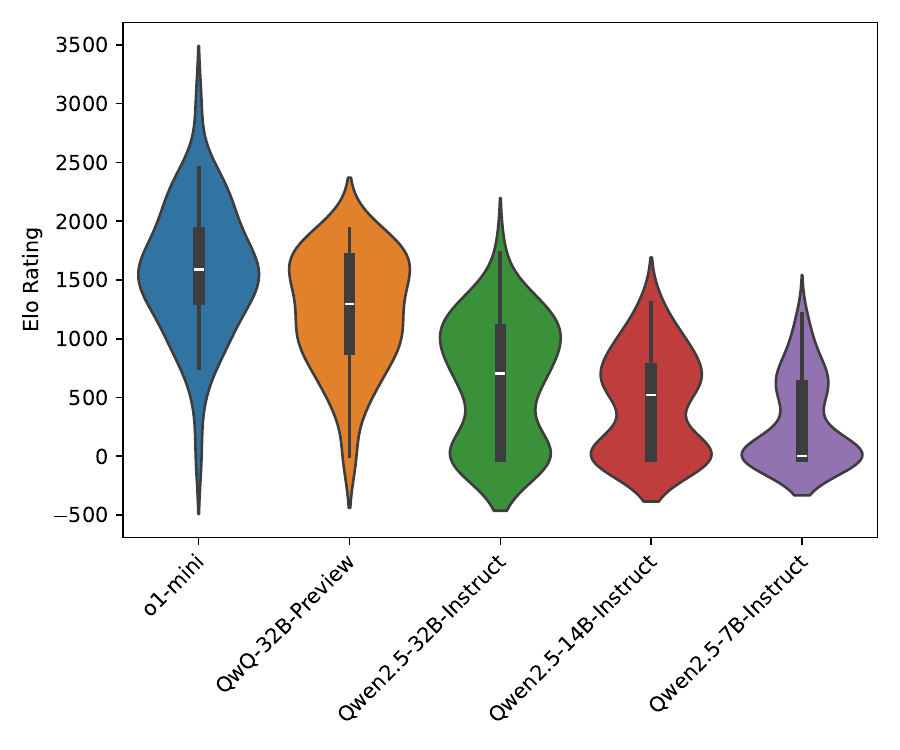}
  \caption{Violin plots of Elo ratings across tested contests.}
  \label{fig:rating-violin}
\end{figure}

The variance in ratings is a crucial aspect of the benchmark. In \Cref{fig:rating-violin}, we present violin plots of several models across all tested contests. We observe that most models exhibit a standard deviation between 300 and 500. This fluctuation in ratings can be attributed to the models' limited ability; solving just one additional problem significantly boosts their ratings in one contest since they can only pass very few. To reduce this standard deviation, increasing the number of tested contests can be beneficial. In our experiments, by testing across 54 contests, the standard deviation in overall average ratings is reduced to around 50, which we is acceptable, and the violin plots can effectively demonstrate their performance comparisons.

\section{Discussion}

\subsection{Contributions}
\label{sec:contributions}

Due to the unique nature of our work compared to other competition code benchmarks, it is important to clearly outline the contributions we make to the NLP community.

\begin{enumerate}[leftmargin=*]

\item Similar to other benchmarks, we provide high-quality test problems. However, unlike others that also include CodeForces problems, our dataset offers a full set of problems that can be updated online and includes detailed information such as contest divisions, problem difficulty ratings, and problem tags. This allows for more comprehensive evaluations compared to existing benchmarks.

\item Our approach stands out by using a unique evaluation method in which problems are submitted directly to the platform. In our previous analysis, we found that this method is crucial for competition code problems, as many of them require special judges and have unaccessible hidden test cases. Unlike general code generation tasks, program efficiency is a critical metric in judging competition code problems. Evaluating them independently can lead to misalignment of the environment and inaccurate results. Our method successfully provides an easy way to address these issues, providing more accurate evaluations that reflect the models' true capabilities.

\item We are the first to provide human-comparable Elo ratings for existing models. This allows us to assess the progress of current AI models in relation to human performance. Our findings suggest that o1-like reasoning models show promise in improving coding capabilities, given their significant advantage over other advanced models on our benchmark. We also provide detailed results across different contest divisions, problem difficulties, and algorithm tags, which can help developers specifically target and improve corresponding abilities.

\item We offer insights that previous work has overlooked or failed to achieve. For example, we find that while Python is the most familiar language for existing LLMs, they often perform better when answering in C++ under \bench{}. This highlights a limitation of previous competition code benchmarks that evaluate solely in Python, as they may not fully stimulate the models' best performance. It also suggests that model developers should consider training their models to output C++ code when efficiency is a critical factor.

\end{enumerate}

\subsection{Limitations}

We believe our benchmark has the following two limitations:

\begin{itemize}[leftmargin=*]
\item One limitation is that we only allow eight submissions per problem. In practice, users can make additional submissions as long as the penalty scores remain lower than the passing scores. This constraint might result in the tested Elo ratings being slightly lower than the actual ratings, as models may successfully solve problems with more attempts. However, it is a carefully considered value to balance the alignment of actual performance while avoiding excessive submissions that could contaminate the platform and impact other users' access, since the margin of larger submission times, like 32 tries or more, will decrease quickly. Therefore, we adopt this setting and encourage others to remain the same for a fair comparison.

\item Another limitation is that we rely on interaction with the CodeForces platform to conduct the judging process, whereas previous benchmarks typically only required offline testing. This reliance is necessary due to the need for bypassing the access to hidden test cases and special judges to provide accurate feedback. In other words, if we had all the hidden test cases and special judges in our hands, we could conduct the evaluation independently. However, these resources are hidden from the platform and extremely difficult to access, and we cannot provide them either.

\end{itemize}

\section{Conclusion}


In this work, we propose \bench{} benchmark, a collection of Codeforces problems with detailed problem information and a unique judgment method that involves submitting solutions to the CodeForces platform and receiving judgment status to achieve zero false positives, special judge support, and an aligned execution environment for the first time. Based on the judgment from the platform, we developed an Elo rating calculation system that aligns with the platform but has a lower variance. Testing on 30 open-source and 3 proprietary LLMs, we find that o1-mini and QwQ-32B-Preview stand out significantly among all models, and most models struggle even with the easiest problems and fall in the lowest 20 percent among all human participants. We further conduct analysis experiments and find different performances of LLMs across algorithm tags, and the best performance is in C++ rather than Python, contradictory with former benchmarks. We have made our \bench{} benchmark publicly available and hope it can pave the way for the NLP community to test LLMs' sophisticated reasoning abilities on code and provide insights for future studies.

\section{Ethical Statement}

Our benchmark relies on the CodeForces platform to conduct judgments. We strictly adhere to the Codeforces Terms and Conditions\footnote{\url{https://codeforces.com/terms}} throughout all experiments and emphasize that others should follow the same. This benchmark is for academic purposes only and should be used in a limited way to avoid impacting user access to the platform. It is designed for virtual participation only and cannot be used for in-contest testing, in accordance with the CodeForces Rule Restricting the Use of AI\footnote{\url{https://codeforces.com/blog/entry/133941}}. Due to ethical considerations, we will conduct a comprehensive risk assessment and seek permission from the CodeForces platform before open-sourcing the entire submission and evaluation scaffold, and we have not included it in this version of the paper. Before that, we recommend that others independently reproduce our proposed method to conduct evaluations. We would also like to express our great acknowledgment to Mike Mirzayanov for creating the remarkable CodeForces platform.







\newpage
\bibliographystyle{plainnat}
\bibliography{custom}

\appendix



\newpage

\section{Model Cards}
\label{appx:model-card}

We list and cite all the tested models in \Cref{tab:model-cards}.

\begin{table*}[!htbp]
    \centering
    \begin{tabular}{l|l|l}
        \toprule
        \textbf{Short Name} & \textbf{Citation} & \textbf{HuggingFace Endpoint} \\ \midrule
        Claude-3-5-Sonnet-2024-10-22 & \citet{claude-3-5} & - \\
        ChatGPT-4o-latest-2024-11-20 & \citet{hurst2024gpt} & - \\
        o1-mini & \citet{o1-mini} & - \\ \midrule

        Qwen2.5-Coder-1.5B-Instruct & \citet{hui2024qwen2} & \href{https://huggingface.co/Qwen/Qwen2.5-Coder-1.5B-Instruct}{Qwen/Qwen2.5-Coder-1.5B-Instruct} \\
        Qwen2.5-Coder-3B-Instruct & \citet{hui2024qwen2} & \href{https://huggingface.co/Qwen/Qwen2.5-Coder-3B-Instruct}{Qwen/Qwen2.5-Coder-3B-Instruct}\\
        Qwen2.5-Coder-7B-Instruct & \citet{hui2024qwen2} & \href{https://huggingface.co/Qwen/Qwen2.5-Coder-7B-Instruct}{Qwen/Qwen2.5-Coder-7B-Instruct}\\
        Qwen2.5-Coder-14B-Instruct & \citet{hui2024qwen2} & \href{https://huggingface.co/Qwen/Qwen2.5-Coder-14B-Instruct}{Qwen/Qwen2.5-Coder-14B-Instruct}\\
        Qwen2.5-Coder-32B-Instruct & \citet{hui2024qwen2} & \href{https://huggingface.co/Qwen/Qwen2.5-Coder-32B-Instruct}{Qwen/Qwen2.5-Coder-32B-Instruct}\\
        Qwen2.5-7B-Instruct & \citet{yang2024qwen2} & \href{https://huggingface.co/Qwen/Qwen2.5-7B-Instruct}{Qwen/Qwen2.5-7B-Instruct}\\
        Qwen2.5-14B-Instruct & \citet{yang2024qwen2} & \href{https://huggingface.co/Qwen/Qwen2.5-14B-Instruct}{Qwen/Qwen2.5-14B-Instruct}\\
        Qwen2.5-32B-Instruct & \citet{yang2024qwen2} & \href{https://huggingface.co/Qwen/Qwen2.5-32B-Instruct}{Qwen/Qwen2.5-32B-Instruct}\\
        Qwen2.5-72B-Instruct & \citet{yang2024qwen2} & \href{https://huggingface.co/Qwen/Qwen2.5-72B-Instruct}{Qwen/Qwen2.5-72B-Instruct}\\ 
        QwQ-32B-Preview & \citet{qwq-32b-preview} & \href{https://huggingface.co/Qwen/QwQ-32B-Preview}{Qwen/QwQ-32B-Preview}\\\midrule

        DS-Coder-1.3B-Instruct & \citet{guo2024deepseek} & \href{https://huggingface.co/deepseek-ai/deepseek-coder-1.3b-instruct}{deepseek-ai/deepseek-coder-1.3b-instruct} \\
        DS-Coder-6.7B-Instruct & \citet{guo2024deepseek} & \href{https://huggingface.co/deepseek-ai/deepseek-coder-6.7b-instruct}{deepseek-ai/deepseek-coder-6.7b-instruct} \\
        DS-Coder-33B-Instruct & \citet{guo2024deepseek} & \href{https://huggingface.co/deepseek-ai/deepseek-coder-33b-instruct}{deepseek-ai/deepseek-coder-33b-instruct} \\
        DS-Coder-V2-Lite-Instruct & \citet{zhu2024deepseek} & \href{https://huggingface.co/deepseek-ai/DeepSeek-Coder-V2-Lite-Instruct}{deepseek-ai/DeepSeek-Coder-V2-Lite-Instruct} \\
        DS-Coder-V2-Instruct & \citet{zhu2024deepseek} & \href{https://huggingface.co/deepseek-ai/DeepSeek-Coder-V2-Instruct}{deepseek-ai/DeepSeek-Coder-V2-Instruct} \\
        DS-V2-Lite-Chat & \citet{liu2024deepseek} & \href{https://huggingface.co/deepseek-ai/DeepSeek-V2-Lite-Chat}{deepseek-ai/DeepSeek-V2-Lite-Chat} \\
        DS-V2-Chat & \citet{liu2024deepseek} & \href{https://huggingface.co/deepseek-ai/DeepSeek-V2-Chat}{deepseek-ai/DeepSeek-V2-Chat}\\
        DS-V2.5 & \citet{liu2024deepseek} & \href{https://huggingface.co/deepseek-ai/DeepSeek-V2.5}{deepseek-ai/DeepSeek-V2.5}\\\midrule

        CodeLlama-70B-Instruct & \citet{roziere2023code} & \href{https://huggingface.co/meta-llama/CodeLlama-70b-Instruct-hf}{meta-llama/CodeLlama-70b-Instruct-hf}\\
        Llama-3.1-8B-Instruct & \citet{dubey2024llama} & \href{https://huggingface.co/meta-llama/Llama-3.1-8B-Instruct}{meta-llama/Llama-3.1-8B-Instruct}\\
        Llama-3.1-70B-Instruct & \citet{dubey2024llama} & \href{https://huggingface.co/meta-llama/Llama-3.1-70B-Instruct}{meta-llama/Llama-3.1-70B-Instruct}\\\midrule

        Codestral-22B-v0.1 & \citet{codestral} & \href{https://huggingface.co/mistralai/Codestral-22B-v0.1}{mistralai/Codestral-22B-v0.1} \\
        Mistral-7B-Instruct-v0.2 & \citet{jiang2023mistral} & \href{https://huggingface.co/mistralai/Mistral-7B-Instruct-v0.2}{mistralai/Mistral-7B-Instruct-v0.2} \\
        Ministral-8B-Instruct-2410 & \citet{jiang2023mistral} & \href{https://huggingface.co/mistralai/Ministral-8B-Instruct-2410}{mistralai/Ministral-8B-Instruct-2410} \\
        Mistral-Large-Instruct-2411 & \citet{jiang2023mistral} & \href{https://huggingface.co/mistralai/Mistral-Large-Instruct-2411}{mistralai/Mistral-Large-Instruct-2411} \\
        Mixtral-8x7B-Instruct-v0.1 & \citet{jiang2024mixtral} & \href{https://huggingface.co/mistralai/Mixtral-8x7B-Instruct-v0.1}{mistralai/Mixtral-8x7B-Instruct-v0.1} \\
        Mixtral-8x22B-Instruct-v0.1 & \citet{jiang2024mixtral} & \href{https://huggingface.co/mistralai/Mixtral-8x22B-Instruct-v0.1}{mistralai/Mixtral-8x22B-Instruct-v0.1} \\\midrule

        OpenCoder-8B-Instruct & \citet{huang2024opencoder} & \href{https://huggingface.co/infly/OpenCoder-8B-Instruct}{infly/OpenCoder-8B-Instruct} \\
        Yi-Coder-9B-Chat & \citet{yicoder} & \href{https://huggingface.co/01-ai/Yi-Coder-9B-Chat}{01-ai/Yi-Coder-9B-Chat} \\
        Starcoder2-15B-Instruct-v0.1 & \citet{wei2024selfcodealign} & \href{https://huggingface.co/bigcode/starcoder2-15b-instruct-v0.1}{bigcode/starcoder2-15b-instruct-v0.1} \\\bottomrule
        
    \end{tabular}
    \caption{Model cards.}
    \label{tab:model-cards}
\end{table*}

\section{Decoding Hyperparameters}

All proprietary models use API calls with default parameters. For open-source models, the inference settings are temperature=0.7, top\_p=0.8, top\_k=20, and repetition\_penalty=1.1. The maximum number of output tokens is set to 4,096 for all models, except for QwQ-32B-Preview, which is set to 32,768 tokens.

\section{Analysis of Our Elo Rating Calculation System}
\label{appx:Elo-rating}

In \Cref{sec:Elo-rating}, we present our method for calculating Elo ratings for each contest. Although there are slight differences between our system and the original CodeForces Elo rating calculation, we provide a proof to demonstrate that our ratings are equivalent to the original. For simplicity, we do not consider the differences between divisions here. In fact, the following analysis will always hold under the same divisions, and since all LLMs attend the same set of contests, it will be fair.

For any specific model, after each contest, we calculate the expected rating and then average them. We consider each contest to be independent, and since we acknowledge that the ratings are standardized, we assume the ratings for any specific model under different contests are independent and identically distributed (IID). Let this expected rating be \( r \), and let the variance be \(\text{Var}(r)\). For a total of \( k \) contests, we calculate the average ratings. Since the ratings in all contests for a specific model are IID, we can easily determine that the average will also have an expected value of \( r \) and a variance of \(\frac{\text{Var}(r)}{k}\). It is evident that the calculated ratings will converge as \( k \) approaches infinity.

In the original calculation from the platform, for each individual, a historical rating list is maintained after each contest, denoted as $r_i$, with an initial value of $r_0 = 0$. After calculating the expected rating $E(r_i)$ in the $i$-th contest based on performance, the rating is updated by moving halfway towards the expected rating from the current rating. Namely, the new rating can be calculated using the formula $r_i = \frac{r_{i-1} + E(r_i)}{2}$. Note that each contest is independent, and $E(r_i)$ shares the same distribution as $r$. Through simple mathematical transformations and deductions, we can determine that the expected value of $r_i$ will converge to $r$, and its variance will converge to \(\text{Var}(r)\) as the number of contests approaches infinity. These results indicate that we can achieve the same expected results as CodeForces while significantly reducing the variance by increasing the number of contests.

\section{Human-comparable Elo Rating}

One advantage of our benchmark is that we provide standardized Elo ratings that are comparable with those of human participants. We present each percentile of ratings among all human participants in \Cref{tab:percentile-rating}, based on publicly available user ratings from the CodeForces platform.

\begin{table*}[ht]
    \centering
    \begin{tabular}{|c|c||c|c||c|c||c|c|}
        \toprule
        Percentile & Rating & Percentile & Rating & Percentile & Rating & Percentile & Rating \\
        \midrule
        1 & 348 & 26 & 740 & 51 & 1088 & 76 & 1390 \\
        2 & 351 & 27 & 754 & 52 & 1103 & 77 & 1398 \\
        3 & 353 & 28 & 767 & 53 & 1118 & 78 & 1405 \\
        4 & 356 & 29 & 781 & 54 & 1133 & 79 & 1411 \\
        5 & 359 & 30 & 794 & 55 & 1147 & 80 & 1418 \\
        6 & 362 & 31 & 806 & 56 & 1162 & 81 & 1427 \\
        7 & 366 & 32 & 819 & 57 & 1176 & 82 & 1437 \\
        8 & 371 & 33 & 832 & 58 & 1191 & 83 & 1448 \\
        9 & 376 & 34 & 845 & 59 & 1205 & 84 & 1462 \\
        10 & 383 & 35 & 858 & 60 & 1218 & 85 & 1478 \\
        11 & 391 & 36 & 872 & 61 & 1231 & 86 & 1497 \\
        12 & 401 & 37 & 886 & 62 & 1243 & 87 & 1518 \\
        13 & 415 & 38 & 900 & 63 & 1254 & 88 & 1543 \\
        14 & 437 & 39 & 913 & 64 & 1265 & 89 & 1571 \\
        15 & 478 & 40 & 927 & 65 & 1276 & 90 & 1603 \\
        16 & 559 & 41 & 942 & 66 & 1288 & 91 & 1624 \\
        17 & 577 & 42 & 956 & 67 & 1301 & 92 & 1648 \\
        18 & 591 & 43 & 971 & 68 & 1313 & 93 & 1678 \\
        19 & 605 & 44 & 985 & 69 & 1325 & 94 & 1712 \\
        20 & 621 & 45 & 1000 & 70 & 1338 & 95 & 1751 \\
        21 & 639 & 46 & 1014 & 71 & 1352 & 96 & 1812 \\
        22 & 662 & 47 & 1029 & 72 & 1365 & 97 & 1916 \\
        23 & 687 & 48 & 1044 & 73 & 1370 & 98 & 2019 \\
        24 & 707 & 49 & 1058 & 74 & 1375 & 99 & 2157 \\
        25 & 724 & 50 & 1073 & 75 & 1382 & 100 & 4009 \\
        \bottomrule
    \end{tabular}
    \caption{Percentiles of ratings among all human participants, calculated based on publicly available user ratings from the CodeForces platform, collected in November, 2024.}
    \label{tab:percentile-rating}
\end{table*}

\newpage
\section{Problem Demonstration}

\begin{figure*}[!h]
  \centering
  \includegraphics[width=\textwidth]{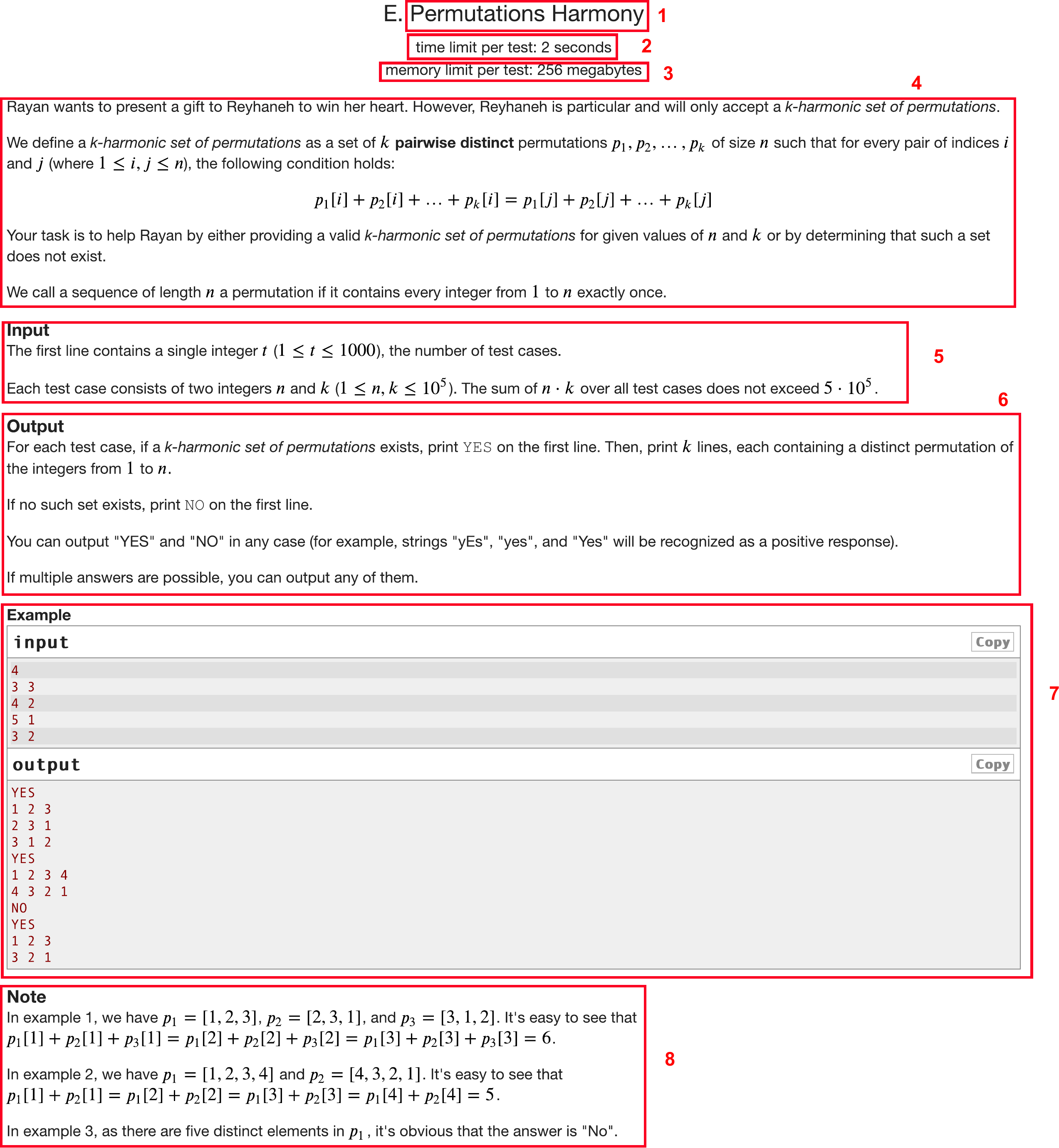}
  \caption{An example of a problem in \bench{}. Each problem contains: 1) title, 2) time limit, 3) memory limit, 4) problem description, 5) input format, 6) output format, 7) test case examples, and 8) note (optional). This problem can be found at \url{https://codeforces.com/contest/2034/problem/E}.}
  \label{fig:prob-demo}
\end{figure*}

A problem demonstration in \bench{} is shown in \Cref{fig:prob-demo}.

\newpage
\section{Special Judge}
\label{appx:spj}

\begin{figure*}[!h]
  \centering
  \includegraphics[width=\textwidth]{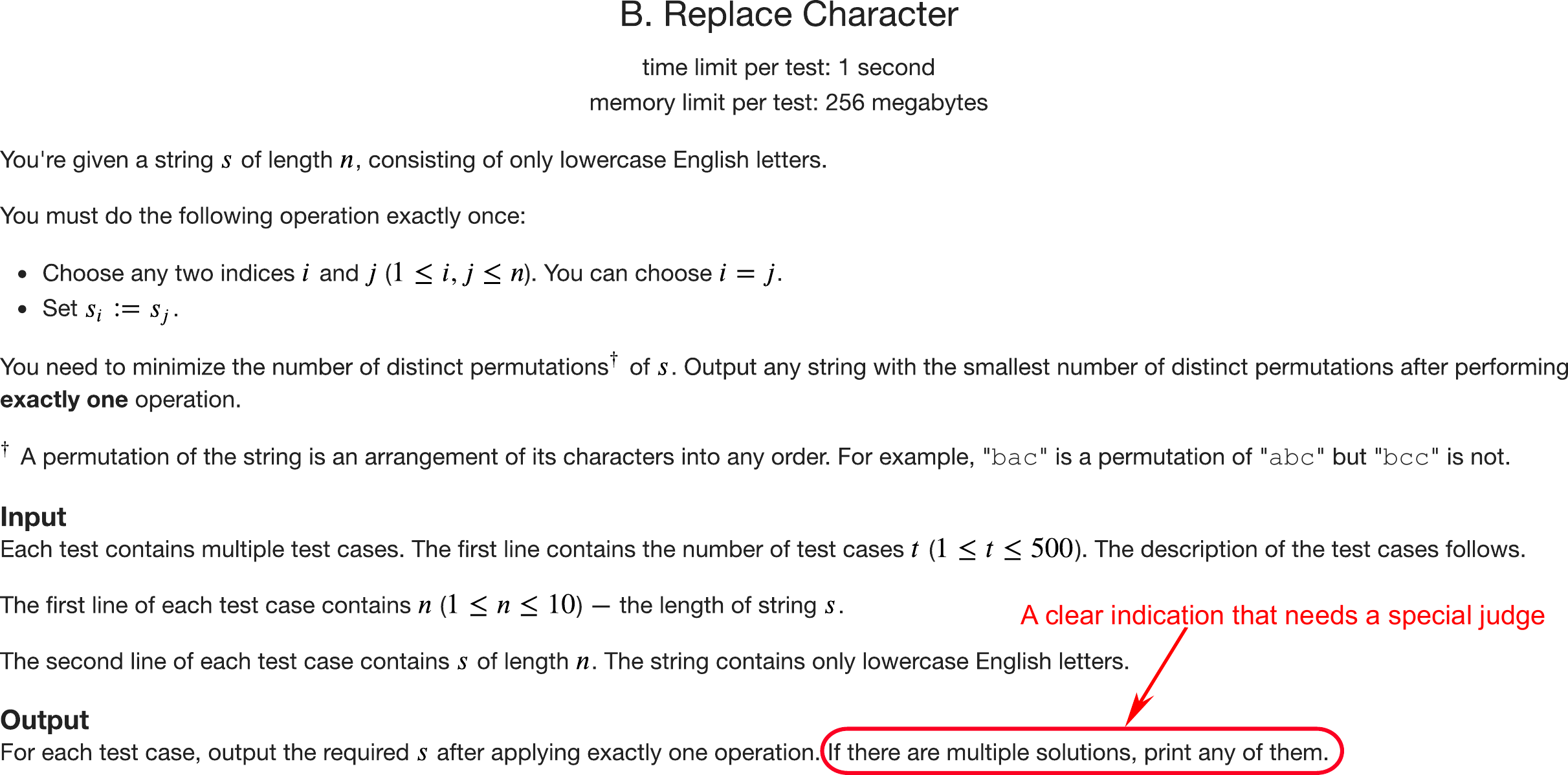}
  \caption{An example of a problem (examples and note parts are omitted) that needs a special judge since there can be multiple valid outputs for the same input (input and outputs refer to test cases but not problem and solutions). \textit{e.g.}, given the input "abc", acceptable outputs could include "abb", "acc", "aac", and any other string derived from "abc" except itself. So we cannot simply compare the output with a predetermined correct solution for evaluation in this problem. \bench{} addresses this by evaluating the code submissions directly on their official platform, marking its first support for this kind of problem. The complete problem can be found at \url{https://codeforces.com/contest/2047/problem/B}.}
  \label{fig:spj-demo}
\end{figure*}

In some code competition problems, a given input might have multiple valid outputs, all of which can be considered correct (Note that the input and outputs here refer to test cases, not the problem statement and model responses). In such situations, a dedicated code is necessary to verify the validity of the outputs instead of simply comparing them against a reference output; this is known as a special judge. It's like a logical unit test, but since the problems are more complex, creating a special judge is also more challenging. \Cref{fig:spj-demo} showcases a case demonstration.

While most competition problems have a single correct output for any given input and do not need a special judge, there are still a proportion of problems that require one. We conducted an empirical study and found that 30 out of 100 randomly selected problems required special judges. Previous competition-level code benchmarks could not handle these situations and therefore did not accurately assess the full capabilities of models. Our evaluation method has the advantage of accommodating these types of problems. Similarly, we also support interactive problems\footnote{An example of interactive problems can be found at \url{https://codeforces.com/contest/2036/problem/G}.} that were not supported in earlier benchmarks. Supporting these kinds of problems is crucial for thoroughly evaluating a model's abilities and obtaining human-comparable Elo ratings.

\end{document}